\documentclass[12pt]{l4dc2021} 

\title[LEOC]{LEOC: A Principled Method in Integrating Reinforcement Learning and Classical Control Theory}
\usepackage{times}
\usepackage{svg}

\usepackage{amsmath,amsfonts,bm}

\def\eqref#1{equation~\ref{#1}}

\def\1{\bm{1}}

\DeclareMathAlphabet{\mathsfit}{\encodingdefault}{\sfdefault}{m}{sl}
\SetMathAlphabet{\mathsfit}{bold}{\encodingdefault}{\sfdefault}{bx}{n}

\newcommand{\R}{\mathbb{R}}

\DeclareMathOperator*{\argmin}{arg\,min}

\usepackage{appendix}
\usepackage{hyperref}
\usepackage{url}

\usepackage{multirow}
\usepackage{amsmath}
\usepackage{amssymb}
\usepackage{graphicx}
\usepackage{algorithm}
\usepackage[noend]{algpseudocode}
\usepackage{wrapfig}
\usepackage{booktabs}
\usepackage{tikz}
\usepackage{textcomp, gensymb}
\usepackage[utf8]{inputenc}
\usepackage[english]{babel}
\usepackage{graphicx}

\author{%
 \Name{Naifu Zhang} \Email{nz248@cantab.net}\\
 \addr Department of Computer Science and Technology, Tsinghua University, Beijing, China
 \AND
 \Name{Nicholas Capel} \Email{nccapel@amazon.com}\\
 \addr Amazon Studios, North Building, 1620 26th St, Santa Monica, USA%
}

\begin{document}

\maketitle

\begin{abstract}%
 There have been attempts in reinforcement learning to exploit \textit{a priori} knowledge about the structure of the system. This paper proposes a hybrid reinforcement learning controller which dynamically interpolates a model-based linear controller and an arbitrary differentiable policy. The linear controller is designed based on local linearised model knowledge, and stabilises the system in a neighbourhood about an operating point. The coefficients of interpolation between the two controllers are determined by a scaled distance function measuring the distance between the current state and the operating point. The overall hybrid controller is proven to maintain the stability guarantee around the neighborhood of the operating point and still possess the universal function approximation property of the arbitrary non-linear policy. Learning has been done on both model-based (PILCO) and model-free (DDPG) frameworks. Simulation experiments performed in OpenAI gym demonstrate stability and robustness of the proposed hybrid controller. This paper thus introduces a principled method allowing for the direct importing of control methodology into reinforcement learning. 
\end{abstract}


\section{Introduction}
\label{sec:introduction}

In recent years, the rise of deep learning has unlocked a new class of function approximation and representation techniques that enable easy discovery of low-dimensional features in extremely high dimensional data \citep{BengioRepresentationPerspectives}. These techniques have naturally been applied to the field of Reinforcement Learning (RL) - a technique that allows agents to learn behaviours through interactions with an environment in order to achieve some desired outcome. 

\begin{wrapfigure}{r}{0.3\textwidth}
  \centering
  \vspace{-20pt}
  \caption{Illustration of operating point in Pendulum}
  \includegraphics[width=0.3\textwidth]{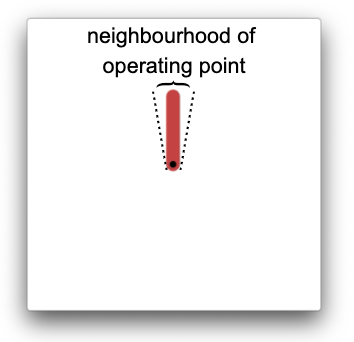}
  \vspace{-40pt}
  \label{fig:operating_point}
\end{wrapfigure}

However, the output policies of the algorithm are opaque, and the relationship between the states and actions cannot be easily understood. Crucially, the policy learnt from optimising a reward function provides no performance guarantee at certain operating regions. For instance, a popular deep RL algorithm Deep Deteriministic Policy Gradient (DDPG) exhibits inconsistent performances \citep{Lillicrap2016ContinuousLearning}, often failing to deliver a truly stable controller \citep{Gu2016Q-Prop:Critic}. When using these techniques for medical purposes \citep{Ferrarese2016MalfunctionsView} or for controlling a chemical plant, control actions with stability and overshoot bounds in an operating point (e.g. a pH or pressure level) could be critical. The ability of users to understand and trust the outputs of these trained controller policies can be key to adoption \citep{Ribeiro2016quotWhyClassifier}.




When we have a system with known local linear dynamics, we can use a powerful suite of tools from control theory to design linear policies with certain highly desired properties, such as better system stability performance in terms of input responses and increased robustness to poor dynamics models. Our \textbf{Locally Enforced Optimal Controller (LEOC)} hybrid policy allows for the manual specification of such a linear controller within a desired region of control, and outside that region, smoothly switches to an arbitrary non-linear policy that is rich enough to model large classes of functions. We theoretically prove that our hybrid policy exhibits the desirable properties of both the linear controller and the learnt arbitrary policy, and verify this experimentally. This paper thus introduces a principled method for controller design that combines tools from control theory with reinforcement learning. 

In this paper, LEOC is utilised within the model-based framework of PILCO \citep{DeisenrothPILCO:Search} and the model-free framework of DDPG \citep{Lillicrap2016ContinuousLearning}, showing that it can in theory be used with other model-based or model-free systems with little to no modification. 


\section{Related Work}
\label{sec:related}

The idea of incorporating existing knowledge into a reinforcement learning model is well-studied. Such techniques include, but are not limited to - Meta Learning \citep{Finn2017Model-AgnosticNetworks,Mishra2017AMeta-Learner,DuanUnderLEARNING} and Hierarchical Reinforcement Learning \citep{NachumData-EfficientLearning,VezhnevetsFeUdalKavukcuoglu,Frans2017MetaHierarchies}, wherein an agent is trained over a distribution of task and learns to quickly adapt its existing algorithms to a new task or learns a meta-policy over subgoals.



The idea of incorporating techniques from control theory into reinforcement learning has also been studied. Techniques such as gain scheduling \citep{Leith2000SurveyDesign}, and iLQG \citep{Todorov2005ASystems} have been employed. However, a major limitation of these techniques is that they require full access to the underlying physics model and its derivatives. In applications, there are many systems that are well studied about certain operating points, with unknown global behaviour outside of them. Rather than embedding prior knowledge into the environment or reward function, we deviate from existing approaches by enforcing the optimal control into the policy itself.

In recent times, the desire to use reinforcement learning in safety-critical settings has inspired some work in formal methods for reinforcement learning. These methods include shielding \citep{Alshiekh2017SafeShielding}, logically constrained learning \citep{Bohrer2018VeriPhy:Models}, justified speculative control \citep{Fulton2018SafeLearning} and constrained Bayesian optimization \citep{GhoshVerifyingOptimization}. These approaches provide formal safety guarantees for reinforcement learning by defining safety specifications, and constraining the learning to a known safe subset of the state space. Yet another subfield of interest is that of robust reinforcement learning, where a policy is considered to be optimal if it has the maximum worst case return \citep{GarcaALearning}. Strands of research involve augmenting the MDP model to create a robust MDP model that provides probabilistic guarantees \citep{Wiesemann2013RobustProcesses}, or by training adversarial agents to provide conflicting inputs to maximize worst case performance \citep{PintoRobustLearning}. 


Our technique is unique in that instead of editing the MDP framework and augmenting it with partial observations, probabilistic guarantees, formal methods or highly parameterised networks, we simply allow for the direct importation of control methodology focused on the local linear approximation about the operating point. This allows for very simple application of well-known control theory, and provides performance guarantees about operating points in which systems spend a majority of their time in.

We hypothesize that this hybrid controller will be very useful in two different kinds of systems: first, systems with general non-linear dynamics, but operating points about which stability is desired; second, systems with very well-studied local behaviour, but unknown global behaviour. Examples include combustion \citep{Fuerhapter2004TheEngine}, continuous non-isothermal stirred tank reactors \citep{Alexandridis2011ControlNetworks}, and catalytic crackers \citep{Arbel1995DynamicsInstabilities}.

\section{Background}
\label{sec:background}

\subsection{RL Set-up}

\begin{wrapfigure}{r}{0.5\textwidth}
  \centering
  \vspace{-40pt}
  \caption{Reinforcement Learning setup}
  \includegraphics[width=0.5\textwidth]{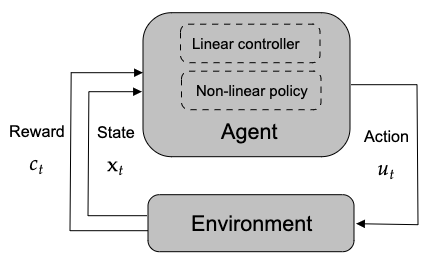}
  \vspace{-40pt}
  \label{fig:setup}
\end{wrapfigure}

We utilize the base reinforcement learning agent and environment setup in \autoref{fig:setup}. Notice that the agent can take any arbitrary non-linear policy. Our model interpolates the outputs of the non-linear controller with a linear controller that takes over at an operating point that is centered at some position in the state space. Our system functions as long as there exists an agent that accepts the state as input and the action as output. 

\subsection{Markov Decision Process (MDP)}


The formal language that we adopt in this paper is one of the Markov Decision Process (MDP) \citep{BellmanAProcess,Puterman1994MarkovianProblems,Ghavamzadeh2016BayesianSurvey} where:
\begin{itemize}
    \item $\mathbf{x}\in X$ to denote a state from a set of measurable, observable and fully controllable possible states
    \item $\mathbf{u} \in U$ to denote an action drawn from a set of possible actions
    \item $P(\mathbf{x}_{t+1}|\mathbf{x}_t,\mathbf{u}_t)\in P(S)$ refers to the transition dynamics, which is the probability of distribution over the next states conditioned on the previous actions and states
    \item $P_0\in P(S)$ denotes the distribution of the initial state
    \item $c(\mathbf{x}, \mathbf{u}) \in P(\mathbb{R})$ represents the reward obtained when an action $\mathbf{u}$ is taken in state $\mathbf{x}$
\end{itemize}

We also introduce a controller or policy function, $\pi(\mathbf{x}_t)$ such that $\mathbf{u}_t=\pi(\mathbf{x}_t)$. The problem of trying to discover the ideal policy $\pi^*(\mathbf{x}_t)$ to achieve the best possible reward thus reduces to:
\begin{equation}
    \pi^*(\mathbf{x}_t) = \argmin_{\pi} \sum_{\forall t} \mathbb{E}[c(\mathbf{x}_t)]
\end{equation}

\subsection{Learning Frameworks}

The optimisation of the baseline controller policies is done in two reinforcement learning frameworks, namely PILCO \citep{DeisenrothPILCO:Search} and DDPG \citep{Lillicrap2016ContinuousLearning}. PILCO belongs to the class of model-based frameworks , whilst DDPG is model-free and is an off-policy algorithm. 

These models are chosen to represent two broad categories of reinforcement learning agents (model-based vs model-free). We hope to show in the following sections that our experiments on these that our hybrid controller generalizes to a wide variety of agents. 

Briefly, a key difference is that PILCO attempts to learn the unknown deterministic transition function $f(\cdot)$, where
\begin{equation}
    \mathbf{x}_{t+1} = f(\mathbf{x}_t, \mathbf{u}_t)
\end{equation}

The probabilistic dynamics model is implemented using a Gaussian process (GP). The GP allows us to perform next-step prediction, propagating state distribution for model evaluation and optimisation.

DDPG, on the other hand, is a model-free algorithm based on the deterministic policy gradient and deep Q-learning that can operate over continuous action spaces. It consists of a critic learned using the Bellman equation as in Q-learning \citep{Watkins1992Q-Learning}, and an actor that is optimised via policy gradient methods \citep{SilverDeterministicAlgorithms}. 


A discussion of these algorithms at length is not within the purview of this paper, and we refer interested reader to the original papers - \cite{Deisenroth2010EfficientProcesses} \cite{DeisenrothPILCO:Search} and \cite{Lillicrap2016ContinuousLearning}.
\section{Overview of Hybrid Controller}
\label{sec:controller}

This section gives a treatment of our hybrid controller policy which consists of a linear and a non-linear part. 

At the operating point, we define a standard linear controller:
\begin{equation}
    G(\mathbf{x})=\mathbf{W}\mathbf{x}+\mathbf{b}
\label{eq:linear_controller}
\end{equation}

where $\mathbf{x} \in \R^D$ is the state, $\mathbf{W} \in \R^{F \times D}$ is a parameter matrix of weights and $\mathbf{b} \in \R^F$ is a bias vector. In each control dimension $d \in [1,...,D]$, the control action is given by a linear combination of states and bias. See \autoref{ssec:linear_controller} and Appendix \ref{appendix:linear_controllers} for more detailed discussions on how to obtain $\mathbf{W}$ and $\mathbf{b}$ in the experiments.

Outside of the neighbourhood of operating point, we switch to a control policy that takes the form of an arbitrary (usually non-linear) function, represented by $H(x)$. 

In the case of PILCO, we choose a radial basis function. In the case of the DDPG, we choose a deep neural network. The requirements of the chosen $H(x)$ is that it is a universal function approximator (i.e. $\exists H$ such that $||H-f||_1<\epsilon$), that it is differentiable with respect to any of the model parameters (i.e. it is trainable), and it accepts the state as an input and outputs the control action.

The overall hybrid policy is thus,

\begin{equation}
    \pi(\mathbf{x}) = r(\mathbf{x}) \underbrace{(\mathbf{W}\mathbf{x}+\mathbf{b})}_{G(\mathbf{x})} + (1-r(\mathbf{x})) H(\mathbf{x})
\label{eq:universal}
\end{equation}


\begin{figure}[h]
  \centering
  \includegraphics[width=0.9\textwidth]{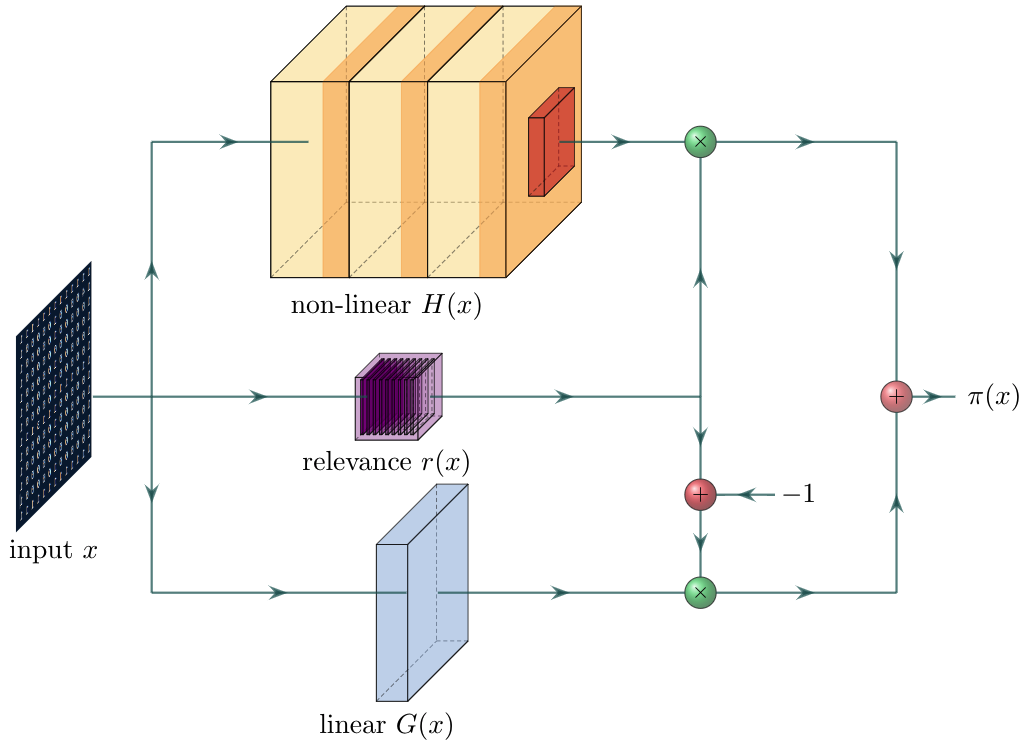}
  \caption{Hybrid controller architecture: $\mathbf{x}$ is passed into the arbitrary non-linear controller $H(\mathbf{x})$, linear controller $G(\mathbf{x})$ and relevance determination function $r(\mathbf{x})$. These three quantities determine the hybrid controller policy $\pi(\mathbf{x})$.}
  \label{fig:internals}
\end{figure}

In the limit that the system is at the operating point, the control action should be that of the linear controller. In general, the closer the system is to the operating point, the more the hybrid controller is to approximate the linear controller. The hybrid controller could therefore be parameterized as a weighted average of the linear and non-linear controllers, with $r(\mathbf{x})$ determining the dependence on either.

\vspace{-20pt}

\begin{subequations}
\begin{align}
\label{eq:R}
r(\mathbf{x}) &= \frac{1}{(1+d(\mathbf{x}))^2}\\
\label{eq:d}
d(\mathbf{x}) &= (\mathbf{x}-\mathbf{a})^{T} \mathbf{\Lambda}^{-1} (\mathbf{x}-\mathbf{a}) \\
\mathbf{\Lambda} &= diag(\lambda_i)
\label{eq:lambda}
\end{align}
\end{subequations}

Naturally, $d(\mathbf{x})$ (\autoref{eq:d}) can be interpreted as a Euclidean distance weighted in different dimensions by $\frac{1}{\lambda_i}$, where $\lambda_i > 0$. With $d(\mathbf{x})$ being computed between the current state $\mathbf{x}$ and the operating point $\mathbf{a}$.


\subsection{Stability Analysis}
\label{ssec:stability}

To prove that stability is maintained, first we consider the linearization of the hybrid controller
\begin{align}
    \pi(\mathbf{x}) &= \pi(\mathbf{a})+ \nabla \pi(\mathbf{a})(\mathbf{x}-\mathbf{a}) + \cdots
\end{align}

In order to evaluate the linearization of $\pi(x)$:
\begin{subequations}
\begin{align}
    \pi(\mathbf{x}) &= r(\mathbf{x})G(\mathbf{x})+(1-r(\mathbf{x}))H(\mathbf{x}) \\
    \nabla \pi(\mathbf{x}) &= \nabla r(\mathbf{x})G(\mathbf{x}) + r(\mathbf{x})\nabla G(\mathbf{x}) + (1-r(\mathbf{x}))\nabla H(\mathbf{x})-\nabla r(\mathbf{x})H(\mathbf{x})
\end{align}
\end{subequations}

When $\mathbf{x}=\mathbf{a}$, $r(\mathbf{a})=1$ and $\nabla r(\mathbf{a})=0$, we can substitute $\mathbf{x}$ with $\mathbf{a}$, obtaining the following
\begin{subequations}
\begin{align}
    \pi(\mathbf{a}) &= G(\mathbf{a}) \\
    \nabla \pi(\mathbf{a}) &= \nabla G(\mathbf{a})
\end{align}
\label{eq:hybrid_same_linear}
\end{subequations}

\autoref{eq:hybrid_same_linear} shows that the local linearization of the hybrid policy reduces to the linear controller $G(\mathbf{a})$. Therefore, if the linear controller is stable about the operating point, the hybrid controller is also stable about the operating point.


\subsection{Proof of Universal Function Approximation}
\label{sec: universalfunction}
We outline only a simplified version of the proof in this section. Please refer to Appendix \ref{appendix:universalproof} for the full proof.

With $r(\mathbf{x})$ defined in \autoref{eq:R}, where $\lambda_i > 0$, $\mathbf{a}\in \R^D$, $\mathbf{W} \in \R^{F\times D}$, $\mathbf{b} \in \R^D$, and the other parameters as defined in \autoref{eq:universal}, in the limit as $\lambda_i \rightarrow \infty$, $r(\mathbf{x})\rightarrow 0$ and $\pi(\mathbf{x})\rightarrow H(\mathbf{x})$. Hence, as long as $H(\mathbf{x})$ is a universal function approximator(i.e. $\exists H$ such that $||H-f||_1<\epsilon$), $\pi(\mathbf{x})$ will also be as well as it learns an appropriate parameterization. The hybrid controller therefore retains its ability to approximate functions in the $\mathbb{L}^1$ space to arbitrary accuracy by learning the right parameterization. 


\section{Experiments}
\label{sec:experiments}

We evaluate our proposed approach on three RL tasks: Swing-up Pendulum, CartPole, and MountainCar. 

The $c(\mathbf{x}, \mathbf{u})$ in each of the environments take the form of the squared distance function, i.e. $c(\mathbf{x}, \mathbf{u}) = - (\mathbf{x}-\mathbf{a})^T\mathbf{K}(\mathbf{x}-\mathbf{a})-k\mathbf{u}^T\mathbf{u}$. For all our hybrid controller experiments we initialize $\mathbf{\Lambda}$ to be the identity matrix. Optimisers used are Adam and BFGS respectively for DDPG and PILCO. For more of the most relevant settings and hyperparameters, refer to Appendix \ref{appendix:hyperparams}.

\subsection{Environments}
\label{sec:environments}


\begin{figure}[h]
  \centering
  \includegraphics[width=\linewidth]{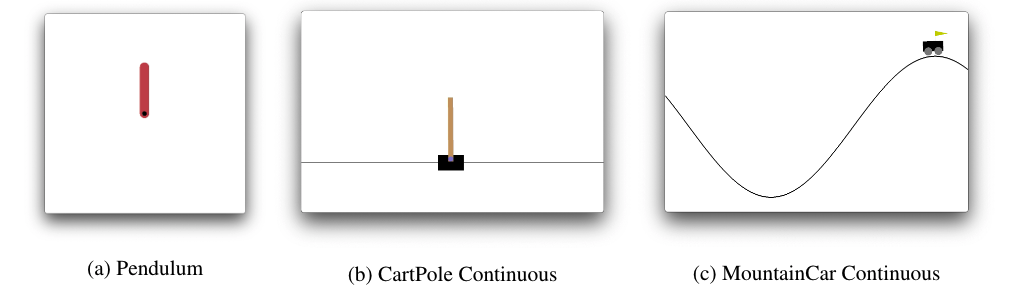}
  \caption{Experimental environments.}
  \label{fig:envs}
  \vspace{-10pt}
\end{figure}

In each of the environments, the agent is initialised off the operating point and the objective is to effect a continuous force on the agent such that it stabilises at the operating point. 


\textbf{Swing-up Pendulum}
OpenAI's Pendulum-v0 (\autoref{fig:envs}a) \citep{Brockman2016OpenAIGym} is used to implement this classical control task. The objective is to swing the pendulum up and balance it in the state $(\theta, \dot{\theta}) = (0, 0)$.

\textbf{CartPole}
OpenAI's CartPole-v1 (\autoref{fig:envs}b) \citep{Brockman2016OpenAIGym} provides an environment consisting of a cart running on a track and a freely swinging pendulum attached to the cart. A non-discrete force $u$ is applied to the cart. The objective is to swing the pendulum up in the middle of the track and balance it in the state $(x, \dot{x}, \theta, \dot{\theta}) = (0, 0, 0, 0)$.


\textbf{MountainCar}
We base the mountain car (\autoref{fig:envs}c) off OpenAI's MountainCarContinuous-v0 environment \citep{Brockman2016OpenAIGym}. In our modification, the cart is rewarded not only for reaching but also staying at the top of the mountain, i.e. $(x, \dot{x}) = (0, 0)$.



\subsection{Linear Controller}
\label{ssec:linear_controller}

For the linear controller design in each of the systems, we defer to the methods developed in optimal control theory.

We derive the optimal state-feedback control gains to achieve closed-loop stable and high performance controller. The gain matrix $\mathbf{K}$ is computed via the Linear Quadratic Regulator (LQR). The feedback action could thus be synthesised by the following
\begin{equation}
\mathbf{u} = -\mathbf{K}\mathbf{x}
\label{eq:linear_controller2}
\end{equation}

\autoref{eq:linear_controller2} is simply \autoref{eq:linear_controller} with $\mathbf{W}=\mathbf{-K}$ and $\mathbf{b}=0$.

Appendix \ref{appendix:linear_controllers} contains the exact form of the local dynamics model in each environment.

\section{Results and Discussion}
\label{sec:results}

\begin{wrapfigure}{r}{0.45\textwidth}
  \centering
  \vspace{-15pt}
  \caption{The dynamics of a trained Pendulum hybrid controller are shown. The blue line plots $\theta$ whereas the orange line plots relevance $r(x)$.}
  \includegraphics[width=0.45\textwidth]{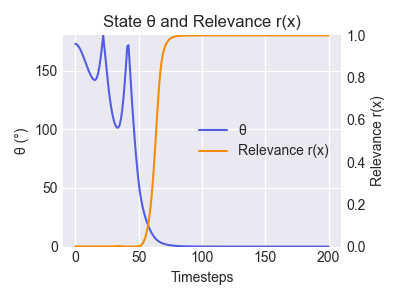}
  \vspace{-50pt}
  \label{fig:theta}
\end{wrapfigure}

In this section, we focus on providing experimental verification some of the theoretical statements we made earlier. In particular, we focus on substantiating the claims in \autoref{ssec:stability} that the LEOC hybrid controller behaves very much like a linear controller around the operating point, thereby also inheriting the desirable properties of the linear controller, namely, \textbf{stability}, \textbf{transient responses}, and \textbf{robustness}.

\subsection{Training and General Comments}

But first, we make a few general comments.

\autoref{fig:theta} illustrates the internals of a trained hybrid controller in the Swing-up Pendulum environment. We can see the linear controller increasingly taking over when $\theta$ gets closer to the operating region. The hybrid controller is hence correctly interpolating between the linear and non-linear policies as designed in \autoref{sec:controller}.

\begin{figure}[h]
  \centering
  \includegraphics[width=\linewidth]{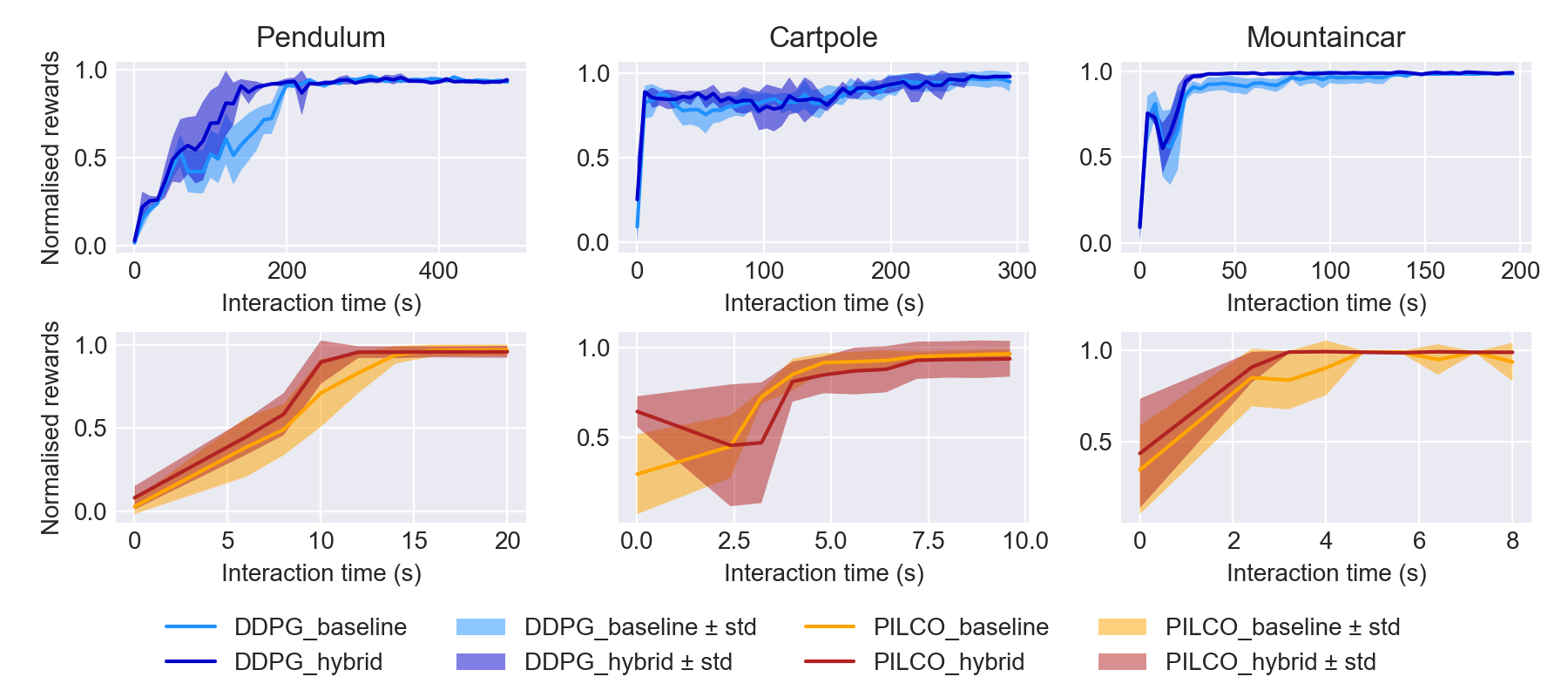}
  \caption{Rewards against interaction time taken to train. The top row compares DDPG baseline with LEOC hybrid implemented in the DDPG framework; the bottom row compares PILCO baseline with LEOC hybrid implemented in PILCO. PILCO is a vastly more data efficient approach than DDPG, and hence takes shorter interaction time.}
  \label{fig:interaction}
\end{figure}

With respect to \autoref{fig:interaction}, we make a couple of observations. First, trivially, convergence in awards can be achieved in training LEOC hybrid controller. This supports the theory in \autoref{sec: universalfunction} that the hybrid controller could approximate the control function as well as the baselines. Second, LEOC takes no longer to train than the respective baselines, in terms of total interaction time prior to convergence. We can note that our hybrid controller converges very quickly once the linear part is activated with $r \approx 1$. The baseline controllers, on the other hand, may require extra system interaction time to explicitly learn the dynamics about the operating point, especially in the Pendulum and MountainCar environments. We report no improvement in interaction time in CartPole, possibly because the linear policy is not activated when the cart is far from the centre position even if the pole is balanced. The bottom line is that learning the LEOC hybrid controller incurs no additional training cost.


\subsection{Stability and Transient Response}

\begin{figure}[h]
  \centering
  \includegraphics[width=\linewidth]{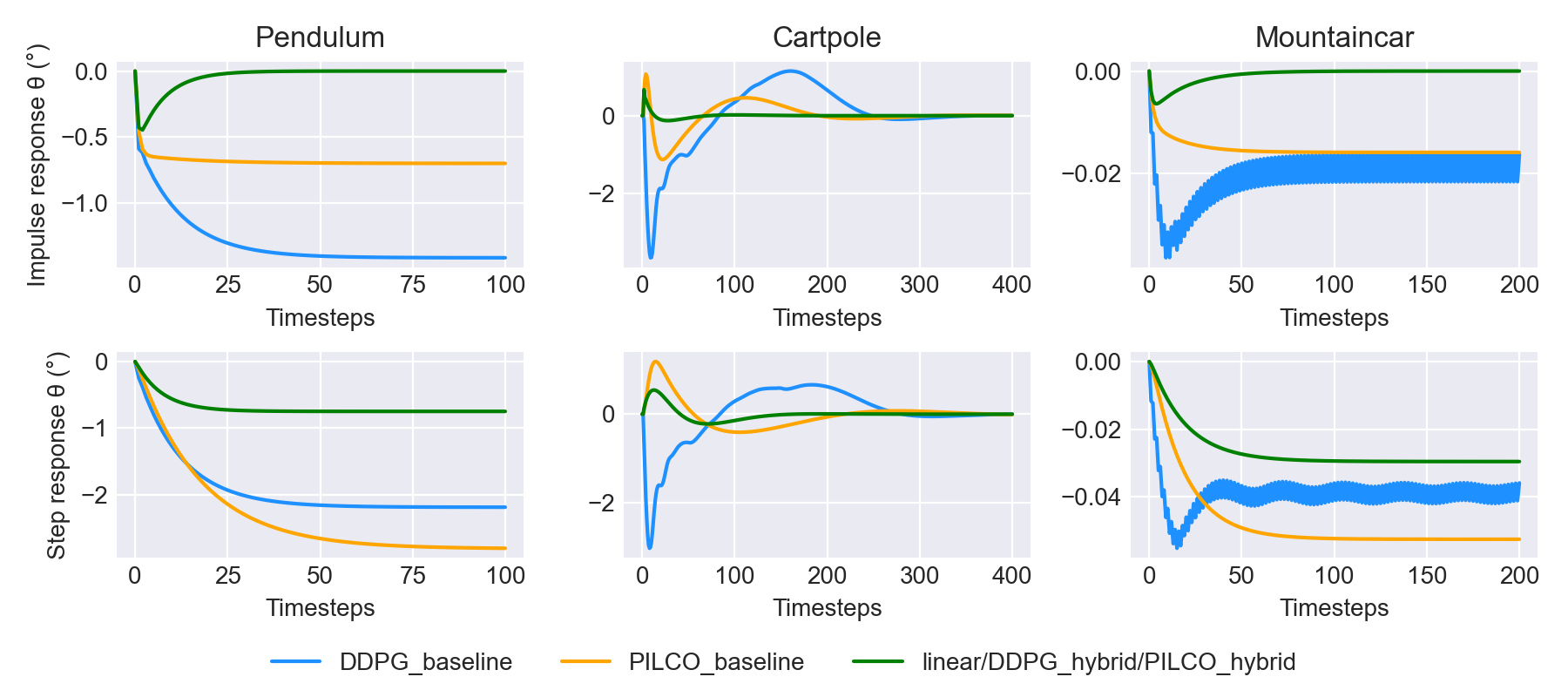}
  \caption{System transient responses in state $\theta$ or $x$, whichever is appropriate for the environment. The top row shows impulse response; the bottom row shows step response. The DDPG\_hybrid and PILCO\_hybrid lines may not be visible as they overlap with the line for linear controller.}
  \label{fig:responses}
\end{figure}

Next, we examine the system's behavior to an impulse and step input when it is at the operating point. \autoref{fig:responses} shows the impulse response and the step response of representative examples of the various policies. 

The impulse response gives insights to the stability and steady state characteristics of the systems. We note that the steady state errors of both DDPG and PILCO baselines are non-zero. There is in fact also no guarantee on the stability of the DDPG and PILCO baselines. As shown in \autoref{fig:responses} and observed by \cite{Gu2016ContinuousAcceleration}, DDPG baseline policy could oscillate around the operating point, therefore not being stable in the Lyapunov stability sense. This serves as empirical evidence to back up the claim made in \autoref{sec:introduction}: that cost-function based approaches in end-to-end reinforcement learning can lead to less desirable properties locally.

The LEOC hybrid controllers, on the other hand, deliver better characteristics in a host of metrics, including steady state errors, peak overshoot and settling time, with smaller means and std, and guaranteed performance (\autoref{table:responses}). The hybrid controllers essentially share the same transient responses as linear controllers, as proven in \autoref{ssec:stability}. This is further evidence that control performance is preserved across the linear and LEOC hybrid controllers at the operating point.

\begin{table}[h]
\centering
\resizebox{\textwidth}{!}{%
\begin{tabular}{llllllll}
                                                   &                                     & \multicolumn{2}{l}{Steady-state Error (\degree)}                                   & \multicolumn{2}{l}{Overshoot (\degree)}                                 & \multicolumn{2}{l}{Settling Time (timesteps)}                        \\ \hline
                                                   &                                     & mean                           & std                           & mean                          & std                          & mean                        & std                       \\ \hline
\multicolumn{1}{|l|}{\multirow{5}{*}{Pendulum}}    & \multicolumn{1}{l|}{DDPG baseline}  & \multicolumn{1}{l|}{-1.79}     & \multicolumn{1}{l|}{0.66}      & \multicolumn{1}{l|}{-}        & \multicolumn{1}{l|}{-}        & \multicolumn{1}{l|}{29.33}  & \multicolumn{1}{l|}{16.13} \\ \cline{2-8} 
\multicolumn{1}{|l|}{}                             & \multicolumn{1}{l|}{DDPG hybrid}    & \multicolumn{1}{l|}{-2.92E-13} & \multicolumn{1}{l|}{0.00}      & \multicolumn{1}{l|}{0.44}     & \multicolumn{1}{l|}{0.00}     & \multicolumn{1}{l|}{26.00}  & \multicolumn{1}{l|}{0.00}  \\ \cline{2-8} 
\multicolumn{1}{|l|}{}                             & \multicolumn{1}{l|}{PILCO baseline} & \multicolumn{1}{l|}{0.33}      & \multicolumn{1}{l|}{6.05}      & \multicolumn{1}{l|}{5.26}     & \multicolumn{1}{l|}{3.33}     & \multicolumn{1}{l|}{49.33}  & \multicolumn{1}{l|}{21.23} \\ \cline{2-8} 
\multicolumn{1}{|l|}{}                             & \multicolumn{1}{l|}{PILCO hybrid}   & \multicolumn{1}{l|}{-3.13E-13} & \multicolumn{1}{l|}{-2.33E-16} & \multicolumn{1}{l|}{0.47}     & \multicolumn{1}{l|}{2.09E-03} & \multicolumn{1}{l|}{26.00}  & \multicolumn{1}{l|}{0.00}  \\ \cline{2-8} 
\multicolumn{1}{|l|}{}                             & \multicolumn{1}{l|}{Linear}         & \multicolumn{1}{l|}{-2.95E-13} & \multicolumn{1}{l|}{0.00}      & \multicolumn{1}{l|}{0.45}     & \multicolumn{1}{l|}{0.00}     & \multicolumn{1}{l|}{26.00}  & \multicolumn{1}{l|}{0.00}  \\ \hline

\end{tabular}%
}
\caption{Impulse response metrics in the Pendulum environment. The LEOC hybrid controllers are essentially the same as linear controllers. We show only one environment here, please refer to Appendix \ref{appendix:metrics} for the full table of all environments.}
\label{table:responses}
\vspace{-20pt}
\end{table}


\subsection{Robustness Analysis}

Finally, we evaluate the robustness and generalization of the learned policy with respect to varying test conditions. We train the policy based on certain parameter (e.g. cart mass or gravitational acceleration constant $g$) values. At test time, however, we evaluate the learned policies by changing parameter values and estimating the average cumulative rewards of a system initialised in the operating region. Specifically, we vary the parameter values from 50\% to 500\% of the original values used in training. 

\begin{figure}[h]
  \centering
  \includegraphics[width=\linewidth]{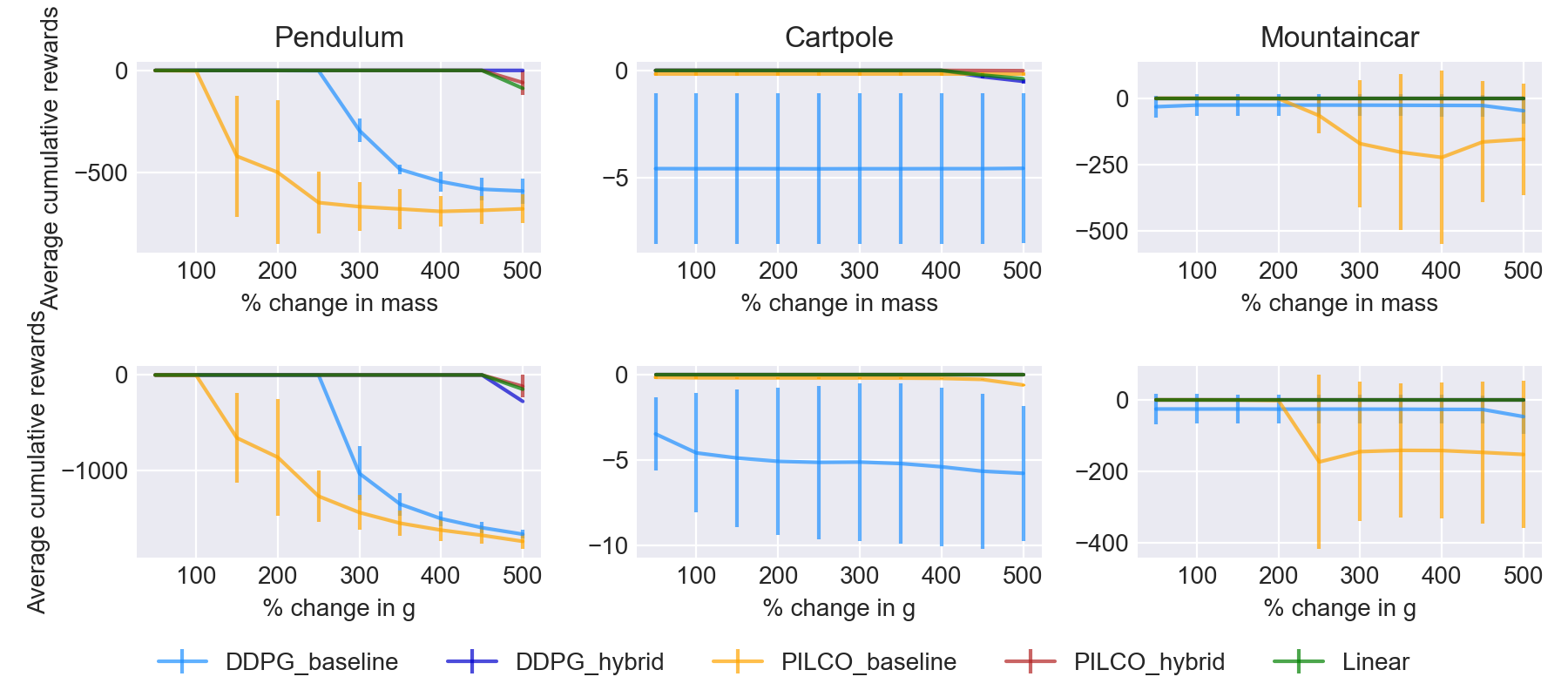}
  \caption{Mean cumulative rewards $\pm$ std against parameter noise. The top row shows the rewards when mass parameters in the environments are changed; the bottom row shows the rewards when $g$ in the environments is changed. The DDPG\_hybrid and PILCO\_hybrid lines may not be too visible as they overlap with the line for linear controller.}
  \label{fig:robustness}
\end{figure}

The rewards are used as a proxy for stability, whereby a unstable system would return a worse reward. As seen in \autoref{fig:robustness}, our LEOC hybrid policies return mostly 0 rewards despite parameter noise and are more robust than either PILCO or DDPG baselines. In fact, LEOC hybrid policies perform as well as linear controllers. 

All in all, we observe that the baseline policies optimised through the traditional reinforcement learning route cannot guarantee stability, transient response performance or robustness. The LEOC hybrid controller, on the other hand, behaves as a linear controller near the operating point, and adopts all the associated desirable properties as theorised in \autoref{ssec:stability}.

\section{Conclusion}
\label{sec:conclusion}
In this paper we introduced LEOC. We proved two desirable properties of our controller - that it is stable about a desired operating point for a given system, and that it retains the ability to serve as a universal function approximator. Via a series of experiments we demonstrate the effectiveness of this new controller by showing similar or better system convergence with lower interaction time. We also demonstrate that it has high robustness to system modelling uncertainties, with performance similar to a pre-designed linear controller, with much better performance than the baseline. We thus demonstrate that our new controller is a practical method for incorporating existing model knowledge into reinforcement learning systems via manipulation of controller models about linearised operating regions. 

Full implementation of the system can be found at \url{https://github.com/nrjc/LEOC}.


\acks{We thank the authors of the GPflow, PILCO and Tensorflow Agents packages for part of our implementation of PILCO and DDPG respectively.}

\bibliography{references}
\clearpage

\appendix

\section{Universal Function Approximation Proof}
\label{appendix:universalproof}
\label{appendix:park_theorem_1}

Theorem 1 presented in \cite{Park1993ApproximationNetworks}, reproduced below, proves that a family of RBF functions is dense in $\mathbb{L}^1(\R^r)$.

\begin{theorem}
Assuming that $K$: $\R^r \rightarrow \R$ is integrable, $S_1(K)$ is dense in $\mathbb{L}^1(\R^r)$ if and only if $\int_{\R^r}K(x)dx\neq 0$.
\end{theorem}

Similar theorems exist for neural networks.

From this fact, we know that $H(x)$ is a universal function approximator. In mathematical terms, that implies that there exists a parameterization of $H(x)$ such that: 
\begin{equation}
    ||H(x)-f||_1<\epsilon 
\label{eq:function_density}
\end{equation}
where $\epsilon \in \R_{>0}$, and $f$ is an arbitrary measurable function. 

\begin{figure}[H]
    \centering
    \includegraphics[width=0.8\textwidth]{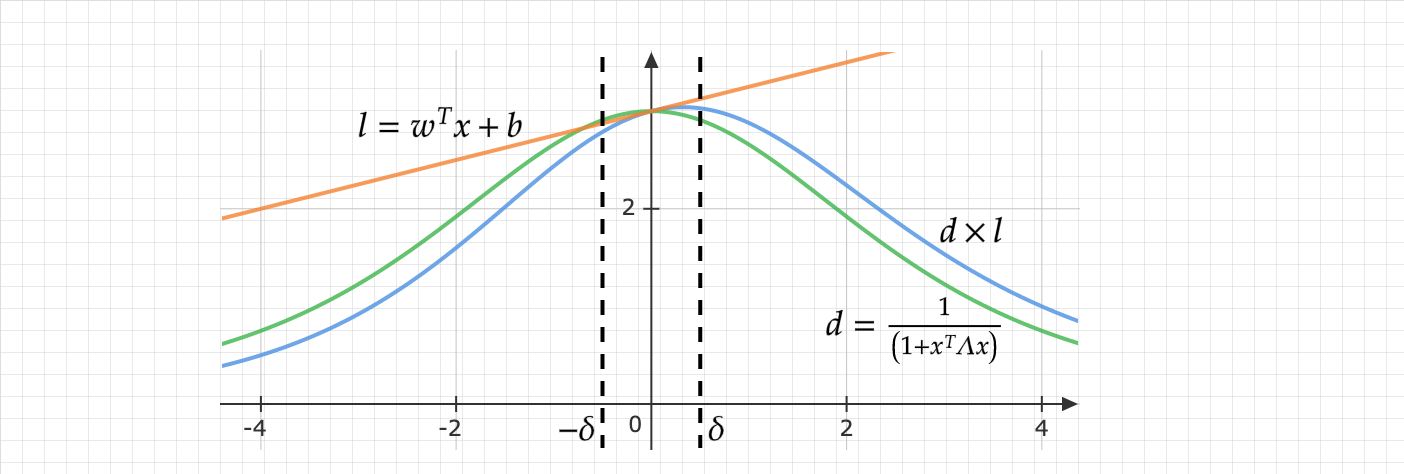}
    \caption{Figure illustrating the parameterization of an arbitrary function in \autoref{appendix:universalproof}}
    \label{fig:epsdelta}
\end{figure}

We will begin our analysis by assuming that $x, c_i\in \R^1$, but the theorem can be trivially extended to arbitrary dimensions, $r$.  \\ 
For any continuous function, $f_c$, there exist some linear function, $y=mx+b$ such that $|f_c-y|<\epsilon$ in an interval $x\in[a-\delta, a+\delta]$, where $\delta \in \R_{>0}$. $H(x)$ can be made arbitrarily small in the domain $x\in[a-\delta, a+\delta]$ due to it being dense in $\mathbb{L}^1$ (Equation \ref{eq:function_density}). With $G(x)=mx+b$, substituting Equation \ref{eq:universal}:

\begin{subequations}
\begin{align}
    |f_c-\pi(x)| &=|f_c-r(x)G(x)-(1-r(x))H(x)|\\
    &\leq |f_c-r(x)G(x)| + |(1-r(x))H(x)|\\
    &\leq |f_c-r(x)G(x)|+ \epsilon/6 \\
    &\leq |f_c - G(x)| + |G(x)-r(x)G(x)| + \epsilon/6
\end{align}
\end{subequations}

In the interval $x\in[a-\delta, a+\delta]$, there exists a $G(x)$ such that $|f_c - G(x)| < \epsilon/6$, as $\delta, \lambda$ can take any non-zero value. Therefore: 

\begin{align}
    |f_c-r(x)G(x)-(1-r(x))H(x)|\leq|G(x)-r(x)G(x)| + \epsilon/3
\end{align}

Consider now:
\begin{subequations}
\begin{align}
    |G(x)-r(x)G(x)| &= |G(x)(1-r(x))| \\
    &\leq ||1-r(x)||_\infty |G(x)|
\end{align}
\end{subequations}

By Holder's inequality. However, $||1-r(x)||_\infty=\frac{\delta^2 \lambda}{1+\delta^2\lambda}$, which can be made arbitrarily small. Therefore, as $|G(x)| < \infty$:

\begin{subequations}
\begin{align}
    |G(x)-r(x)G(x)| &< \epsilon/6 \\
    |f_c-r(x)G(x)-(1-r(x))H(x)|  &< \epsilon/2
\end{align}
\end{subequations}

In the region $x \notin [a-\delta, a+\delta]$:

\begin{subequations}
\begin{align}
|f_c-\pi|&=|f_c-r(x)G(x)-(1-r(x))H(x)|\\
&= |f_e-r'(x)H(x)|\\
&= |f_fr'(x) - r'(x)H(x)|\\
&\leq |r'(x)|_\infty |f_f-H(x)|\\
&=|f_f-H(x)| < \epsilon/2
\end{align}
\end{subequations}

Once again by Holder's inequality, and the fact that $H(x)$ is dense in $\mathbb{L}_1$. Where $r'(x)=1-r(x)$, $f_e = f_c-r(x)G(x) \in \mathbb{L}^1(\R)$, and $f_f = \frac{f_e}{r'(x)} \in \mathbb{L}^1(\R)$ as $r'(x) > \frac{\delta^2 \lambda}{1+\delta^2\lambda}$\\

Thus: 
\begin{equation}
|f_c-\pi|<\epsilon
\end{equation}
for all $x \notin [a-\delta, a+\delta]$. \\

Therefore, $|f_c-\pi| < \epsilon$, $\forall x$ and our family of functions is dense in the space of continuous functions and is thus dense in $L^1$

\clearpage

\section{Linear Controllers in Experiments}
\label{appendix:linear_controllers}

Each of these systems follows the state-space equation in the canonical form, where $\mathbf{x}$ represents the observed states, and $\mathbf{y}$ the output. 

\begin{subequations}
\begin{align}
    \label{eq:statespace}
    \dot{\mathbf{x}} = \mathbf{A}\mathbf{x} + \mathbf{Bu} \\
    \mathbf{y} = \mathbf{C}\mathbf{x} + \mathbf{Du}
\end{align}
\end{subequations}

We describe each of the linear controllers by their state-space equations about the equilibrium point. These local equations are linearly approximated wherever appropriate. All systems are frictionless. All systems follow the canonical state-space equation of 
\autoref{eq:statespace}.

\begin{equation}
\begin{array}{ c c c c c c c }
\overbrace{\textcolor[rgb]{1,1,1}{--}}^{\dot{\mathbf{x}}} & & \overbrace{\textcolor[rgb]{1,1,1}{-----------}}^{\mathbf{A}} & \overbrace{\textcolor[rgb]{1,1,1}{--}}^{\mathbf{x}} & & \overbrace{\textcolor[rgb]{1,1,1}{--------}}^{\mathbf{B}} & \overbrace{\textcolor[rgb]{1,1,1}{-}}^{\mathbf{u}}
\\
\multicolumn{7}{l}{\text{Pendulum}} \\
\begin{pmatrix}
\dot{\theta }\\
\ddot{\theta }
\end{pmatrix} & = & \begin{pmatrix}
0 & 1\\
\frac{mlg}{ml^{2} +I} & 0
\end{pmatrix} & \begin{pmatrix}
\theta \\
\dot{\theta }
\end{pmatrix} & + & \begin{pmatrix}
0\\
-\frac{1}{ml^{2} +I}
\end{pmatrix} & u\\ \\
\multicolumn{7}{l}{\text{CartPole}} \\
\begin{pmatrix}
\dot{x}\\
\ddot{x}\\
\dot{\theta }\\
\ddot{\theta }
\end{pmatrix} & = & \begin{pmatrix}
0 & 1 & 0 & 0\\
0 & 0 & \frac{m^{2} l^{2} g}{I(M+m)+Mml^{2}} & 0\\
0 & 0 & 0 & 1\\
0 & 0 & \frac{mlg(M+m)}{I(M+m)+Mml^{2}} & 0
\end{pmatrix} & \begin{pmatrix}
x\\
\dot{x}\\
\theta \\
\dot{\theta }
\end{pmatrix} & + & \begin{pmatrix}
0\\
-\frac{I+Mml^{2}}{I(M+m)+Mml^{2}}\\
0\\
\frac{ml}{I(M+m)+Mml^{2}}
\end{pmatrix} & u\\ \\
\multicolumn{7}{l}{\text{MountainCar}} \\
\begin{pmatrix}
\dot{x}\\
\ddot{x}
\end{pmatrix} & = & \begin{pmatrix}
0 & 1\\
g & 0
\end{pmatrix} & \begin{pmatrix}
x\\
\dot{x}
\end{pmatrix} & + & \begin{pmatrix}
0\\
-\frac{1}{M}
\end{pmatrix} & u \\
\end{array}
\end{equation}

\clearpage

\section{Hyperparameters}
\label{appendix:hyperparams}

\begin{table}[H]
\centering
\begin{tabular}{llll}
\hline
Env            & CartPole                 & MountainCar              & Pendulum                 \\ \hline
Initialization & [0, 0, -1, 0, 0]             & [$-\pi$,   0]            & [-1,   0,  0]            \\
Target         & [0, 0, 1, 0, 0]          & [0, 0]                   & [1, 0, 0]  \\
Rewards weight $\mathbf{K}$       & diag([1, 0.1]) & diag([0.1, 0, 0.5, 0]) &  diag([0.5, 0.1]) \\
Rewards weight $k$         & 0.001         & 0.005                   & 0.005  
\\ \hline
\end{tabular}
\caption{Environmental hyperparameters}

\end{table}
\begin{table}[H]
\centering
\resizebox{\textwidth/3*2}{!}{%
\begin{tabular}{lll}
\hline
Framework                       & DDPG                  & PILCO \\ \hline
Optimizer                       & AdamOptimiser         & BFGS  \\
Number   of iterations/rollouts & 10000-15000           & 10-12 \\
Actor   Learning Rate           & $5\times10^{-4}$      & -      \\
Critic   Learning Rate          & $1\times10^{-3}$      & -      \\
Learning   rates                & $5\times10^{-4}$      & -      \\
Discount   Factor               & 0.995                 & -      \\
Minibatch   size                & 64                    & -      \\
Memory   Buffer Size            & 100000                & -     \\
Target   update factor          & 0.05                  & -      \\
Target   update rate            & 5                     & -      \\
Randomise controller mean            & -                     & 1      \\
Randomise controller std            & -                     & 0.01      \\
$\Lambda$ in hybrid policy    & Identity              & Identity    \\ \hline
\end{tabular}
}
\caption{Framework hyperparameters}

\end{table}

\clearpage


\section{Impulse Response Metrics}
\label{appendix:metrics}
\begin{table}[h]
\centering
\resizebox{\textheight/3*2}{!}{%
\begin{tabular}{llllllll}
                                                   &                                     & \multicolumn{2}{l}{Steady State Error}                                   & \multicolumn{2}{l}{Overshoot}                                 & \multicolumn{2}{l}{Settling Time}                        \\ \hline
                                                   &                                     & Mean                           & S.D.                           & Mean                          & S.D.                          & Mean                        & S.D.                       \\ \hline
\multicolumn{1}{|l|}{\multirow{5}{*}{Pendulum}}    & \multicolumn{1}{l|}{DDPG baseline}  & \multicolumn{1}{l|}{-1.79}     & \multicolumn{1}{l|}{0.66}      & \multicolumn{1}{l|}{-}        & \multicolumn{1}{l|}{-}        & \multicolumn{1}{l|}{29.33}  & \multicolumn{1}{l|}{16.13} \\ \cline{2-8} 
\multicolumn{1}{|l|}{}                             & \multicolumn{1}{l|}{DDPG hybrid}    & \multicolumn{1}{l|}{-2.92E-13} & \multicolumn{1}{l|}{0.00}      & \multicolumn{1}{l|}{0.44}     & \multicolumn{1}{l|}{0.00}     & \multicolumn{1}{l|}{26.00}  & \multicolumn{1}{l|}{0.00}  \\ \cline{2-8} 
\multicolumn{1}{|l|}{}                             & \multicolumn{1}{l|}{PILCO baseline} & \multicolumn{1}{l|}{0.33}      & \multicolumn{1}{l|}{6.05}      & \multicolumn{1}{l|}{5.26}     & \multicolumn{1}{l|}{3.33}     & \multicolumn{1}{l|}{49.33}  & \multicolumn{1}{l|}{21.23} \\ \cline{2-8} 
\multicolumn{1}{|l|}{}                             & \multicolumn{1}{l|}{PILCO hybrid}   & \multicolumn{1}{l|}{-3.13E-13} & \multicolumn{1}{l|}{-2.33E-16} & \multicolumn{1}{l|}{0.47}     & \multicolumn{1}{l|}{2.09E-03} & \multicolumn{1}{l|}{26.00}  & \multicolumn{1}{l|}{0.00}  \\ \cline{2-8} 
\multicolumn{1}{|l|}{}                             & \multicolumn{1}{l|}{Linear}         & \multicolumn{1}{l|}{-2.95E-13} & \multicolumn{1}{l|}{0.00}      & \multicolumn{1}{l|}{0.45}     & \multicolumn{1}{l|}{0.00}     & \multicolumn{1}{l|}{26.00}  & \multicolumn{1}{l|}{0.00}  \\ \hline
\multicolumn{1}{|l|}{\multirow{5}{*}{CartPole}}    & \multicolumn{1}{l|}{DDPG baseline}  & \multicolumn{1}{l|}{9.97E-04}  & \multicolumn{1}{l|}{3.79E-03}  & \multicolumn{1}{l|}{2.88}     & \multicolumn{1}{l|}{0.86}     & \multicolumn{1}{l|}{282.25} & \multicolumn{1}{l|}{75.40} \\ \cline{2-8} 
\multicolumn{1}{|l|}{}                             & \multicolumn{1}{l|}{DDPG hybrid}    & \multicolumn{1}{l|}{4.79E-06}  & \multicolumn{1}{l|}{7.46E-07}  & \multicolumn{1}{l|}{0.67}     & \multicolumn{1}{l|}{7.46E-07} & \multicolumn{1}{l|}{121.00} & \multicolumn{1}{l|}{5.02}  \\ \cline{2-8} 
\multicolumn{1}{|l|}{}                             & \multicolumn{1}{l|}{PILCO baseline} & \multicolumn{1}{l|}{2.63E-02}  & \multicolumn{1}{l|}{1.41E-02}  & \multicolumn{1}{l|}{1.34}     & \multicolumn{1}{l|}{0.15}     & \multicolumn{1}{l|}{262.67} & \multicolumn{1}{l|}{27.81} \\ \cline{2-8} 
\multicolumn{1}{|l|}{}                             & \multicolumn{1}{l|}{PILCO hybrid}   & \multicolumn{1}{l|}{7.16E-06}  & \multicolumn{1}{l|}{6.69E-07}  & \multicolumn{1}{l|}{0.93}     & \multicolumn{1}{l|}{2.97E-02} & \multicolumn{1}{l|}{138.33} & \multicolumn{1}{l|}{2.62}  \\ \cline{2-8} 
\multicolumn{1}{|l|}{}                             & \multicolumn{1}{l|}{Linear}         & \multicolumn{1}{l|}{4.17E-06}  & \multicolumn{1}{l|}{0.00}      & \multicolumn{1}{l|}{0.67}     & \multicolumn{1}{l|}{0.00}     & \multicolumn{1}{l|}{117.00} & \multicolumn{1}{l|}{0.00}  \\ \hline
\multicolumn{1}{|l|}{\multirow{5}{*}{MountainCar}} & \multicolumn{1}{l|}{DDPG baseline}  & \multicolumn{1}{l|}{0.24}      & \multicolumn{1}{l|}{1.64E-02}  & \multicolumn{1}{l|}{-}        & \multicolumn{1}{l|}{-}        & \multicolumn{1}{l|}{33.67}  & \multicolumn{1}{l|}{1.25}  \\ \cline{2-8} 
\multicolumn{1}{|l|}{}                             & \multicolumn{1}{l|}{DDPG hybrid}    & \multicolumn{1}{l|}{-2.11E-07} & \multicolumn{1}{l|}{1.23E-08}  & \multicolumn{1}{l|}{6.39E-03} & \multicolumn{1}{l|}{4.11E-04} & \multicolumn{1}{l|}{201.50} & \multicolumn{1}{l|}{0.73}  \\ \cline{2-8} 
\multicolumn{1}{|l|}{}                             & \multicolumn{1}{l|}{PILCO baseline} & \multicolumn{1}{l|}{-1.50E-02} & \multicolumn{1}{l|}{1.27E-03}  & \multicolumn{1}{l|}{2.07E-02} & \multicolumn{1}{l|}{6.69E-03} & \multicolumn{1}{l|}{147.00} & \multicolumn{1}{l|}{76.37} \\ \cline{2-8} 
\multicolumn{1}{|l|}{}                             & \multicolumn{1}{l|}{PILCO hybrid}   & \multicolumn{1}{l|}{-2.23E-07} & \multicolumn{1}{l|}{1.25E-09}  & \multicolumn{1}{l|}{6.73E-03} & \multicolumn{1}{l|}{3.46E-05} & \multicolumn{1}{l|}{201.00} & \multicolumn{1}{l|}{0.50}  \\ \cline{2-8} 
\multicolumn{1}{|l|}{}                             & \multicolumn{1}{l|}{Linear}         & \multicolumn{1}{l|}{-2.12E-07} & \multicolumn{1}{l|}{0.00}      & \multicolumn{1}{l|}{6.43E-03} & \multicolumn{1}{l|}{0.00}     & \multicolumn{1}{l|}{201.00} & \multicolumn{1}{l|}{0.00}  \\ \hline
\end{tabular}%
}
\caption{Impulse response metrics}
\end{table}





\end{document}